\pdfoutput=1
\documentclass{bmvc2k}

\title{GRCD: Grounded Region Change Detection for Multi-Finding Chest X-Ray Pairs}

\addauthor{OFM Riaz Rahman Aranya}{ofmriazrahman.aranya@utsa.edu}{}
\addauthor{Peyman Najafirad}{peyman.najafirad@utsa.edu}{}
\addauthor{Kevin Desai}{kevin.desai@utsa.edu}{}

\addinstitution{Department of Computer Science\\The University of Texas at San Antonio \\San Antonio, Texas, USA}

\runninghead{Aranya, Najafirad, Desai}{GRCD}

\def\eg{\emph{e.g}\bmvaOneDot}

\def\etal{\emph{et al}\bmvaOneDot}

\usepackage{booktabs}
\usepackage{amssymb}
\usepackage{tikz}
\usetikzlibrary{arrows.meta, positioning, shapes.geometric, calc, backgrounds}

\makeatletter
\renewcommand\section{\@startsection{section}{1}{\z@}%
  {0.5ex plus .3ex minus .1ex}{0.3ex plus .1ex}%
  {\normalfont\large\bfseries}}
\renewcommand\subsection{\@startsection{subsection}{2}{\z@}%
  {0.3ex plus .2ex minus .1ex}{0.1ex plus .1ex}%
  {\normalfont\normalsize\bfseries}}
\makeatother

\begin{document}

\maketitle

\begin{abstract}
Radiologists routinely compare current and prior chest X-rays to track disease progression, producing follow-up reports that describe multiple findings, each localised to an anatomical region and annotated with a temporal change status. Existing automated methods either generate reports from a single image without modelling temporal context, or incorporate temporal information but do not ground their outputs spatially. The few approaches that combine temporal reasoning with spatial grounding are restricted to single-finding descriptions, leaving multi-finding reports with mixed change directions unaddressed. We present GRCD, a framework for grounded report generation from chest X-ray pairs in the multi-finding setting. We first construct a rigorously cleaned dataset of temporal chest X-ray pairs by identifying and correcting two systematic labelling errors in the source annotations. We then introduce a Region-Guided Change Token module that encodes per-region temporal change across anatomical structures and injects this signal into a language model through a dual-pathway strategy combining prepended spatial tokens with gated cross-attention. On a multi-finding test set, GRCD outperforms existing baselines on text generation and clinical accuracy metrics, with gains in change detection. Ablation studies confirm that the dual-pathway design outperforms either integration strategy in isolation on text and clinical metrics, and that region-level change encoding is necessary for multi-finding generation. Code is available at 
\href{https://github.com/UTSA-VIRLab/GRCD}{https://github.com/UTSA-VIRLab/GRCD}
\end{abstract}

\section{Introduction}
\label{sec:intro}

Chest radiography is the most frequently performed diagnostic
imaging examination worldwide. In routine clinical practice,
radiologists do not interpret chest X-rays (CXRs) in isolation;
they compare each new study against one or more prior images from
the same patient to assess disease progression, treatment
response, or post-procedural status. The resulting follow-up
report describes \emph{multiple} findings, each localised to an
anatomical region and annotated with a temporal change status such
as \emph{worsened}, \emph{improved}, or \emph{stable}. Producing
these reports is time-consuming and cognitively demanding, and the
growing volume of imaging studies has made automated
decision-support tools increasingly desirable.

Existing work on automated CXR report generation falls into 
three groups, each addressing part of this problem but not 
all of it. Single-image methods~\cite{chen2020r2gen, 
liu2024llava, cxrllava2024, pellegrini2023radialog} generate 
fluent reports but cannot describe how findings have changed 
relative to a prior study. Temporal 
methods~\cite{hergen2024, mlrg2025, ddatr2025, libra2025} 
compare current and prior images but do not ground their 
outputs spatially. Grounded methods~\cite{tanida2023rgrg, 
bannur2024maira2, padchestgr2024} localise findings with 
bounding boxes but operate on single time points. 
TRACE~\cite{tracev1} combines temporal reasoning with spatial 
grounding but is restricted to single-finding descriptions, 
whereas real follow-up reports contain multiple findings with 
mixed change directions.

\begin{figure*}[t]
\centering
\includegraphics[width=0.65\textwidth]{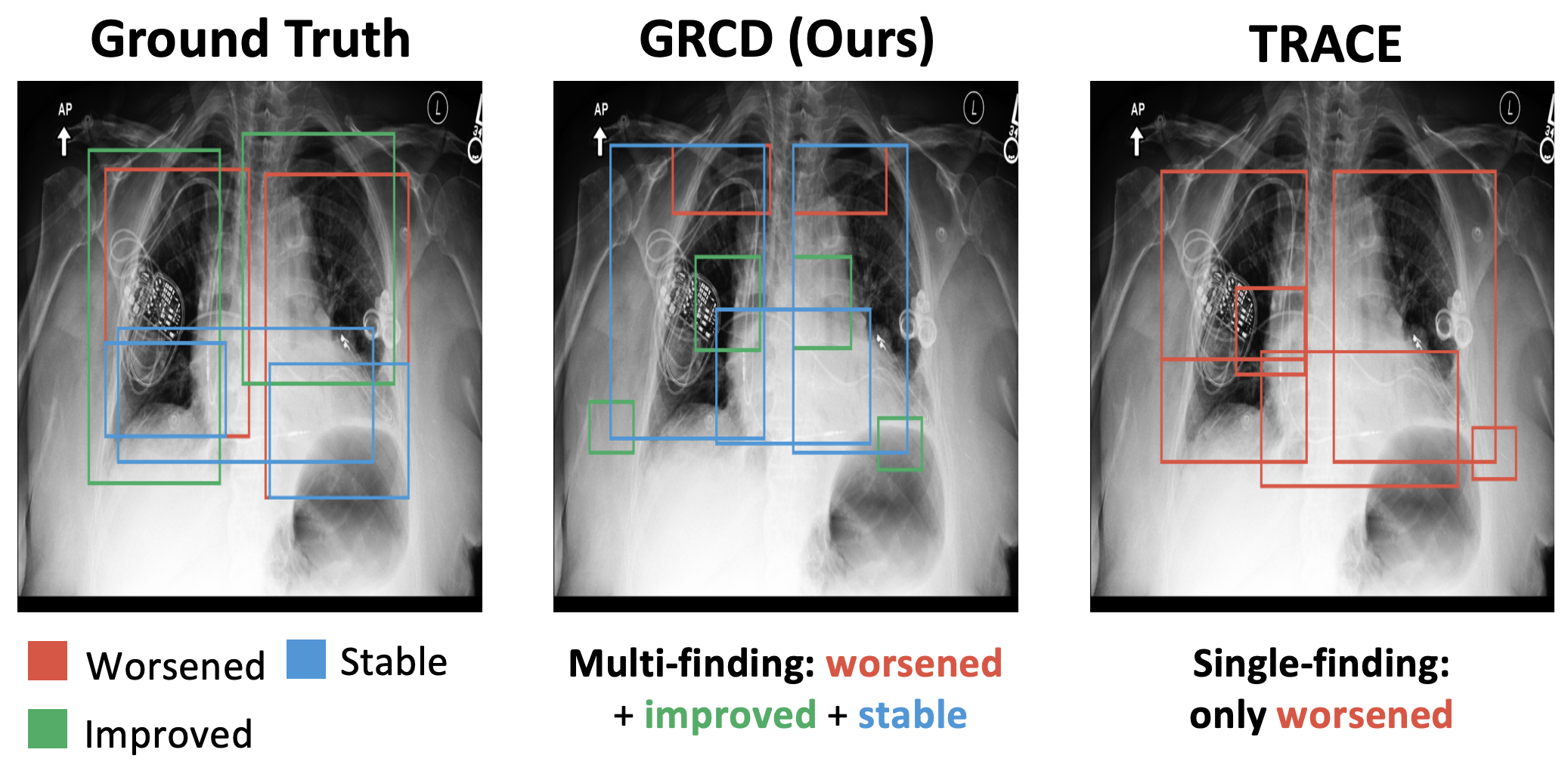}
\caption{GRCD (Ours) generates structured multi-finding reports with distinct change labels (worsened, improved, stable) and grounded bounding boxes, while TRACE~\cite{tracev1} collapses to a single repeated label across all regions.}
\label{fig:motivation}
\end{figure*}

The multi-finding setting introduces challenges absent from 
single-finding generation: the model must track change across 
many anatomical regions simultaneously, resolve overlapping 
bounding boxes, and maintain clinical coherence across a 
variable number of output sentences. We show empirically that 
a model without explicit region-level encoding collapses into 
repetitive output when faced with this task. As illustrated in 
Fig.~\ref{fig:motivation}, TRACE produces only a single 
repeated finding type, while our proposed GRCD correctly 
identifies multiple change types with distinct grounded regions 
within the same study.

In this paper, we present GRCD, a framework for grounded 
temporal report generation that handles the clinically 
realistic multi-finding setting. Our contributions are:

\begin{itemize}
\item The first model to generate multi-finding temporal 
  reports with per-finding spatial grounding and change labels.

\item A three-tier data cleaning pipeline that identifies and 
  corrects two systematic errors in Chest ImaGenome temporal 
  pairs, yielding a 40{,}250-pair benchmark.

\item A Region-Guided Change Token (RGCT) module with dual-pathway integration that we show is necessary for multi-finding generation.
\end{itemize}
\section{Related Work}
\label{sec:related}

\paragraph{Radiology report generation.}
Early work framed the task as image captioning, using CNN 
encoders paired with RNN or Transformer 
decoders~\cite{chen2020r2gen}. The adoption of large 
vision-language models has since improved both fluency and 
clinical accuracy. CXR-LLaVA~\cite{cxrllava2024} and 
RaDialog~\cite{pellegrini2023radialog} fine-tune the LLaVA 
architecture for chest radiograph interpretation, with 
RaDialog adding a structured pathology module for interactive 
report editing. CheXagent~\cite{chen2024chexagent} scales up 
to 8B parameters and trains across 28 CXR-specific tasks 
including report generation, visual question answering, and 
disease classification. Uni-Med~\cite{unimed2024} takes a 
multi-task approach with a shared architecture that handles 
six medical tasks through connector mixture-of-experts. 
LLM-RG4~\cite{llmrg4_2025} accepts diverse input 
configurations including multi-view and longitudinal images 
through adaptive token fusion. While these methods produce 
increasingly fluent reports, none of them explicitly model 
how findings change between a prior and current study.

\paragraph{Temporal CXR analysis.}
In clinical practice, radiologists compare the current study 
with prior images, yet most generation methods condition on 
only a single image. BioViL-T~\cite{bannur2023biovilt} 
introduced a CNN-Transformer hybrid pre-trained with a 
temporal contrastive objective that captures progression 
patterns, though it does not generate full reports. Among 
generation methods, HERGen~\cite{hergen2024} integrates up to 
five prior studies through a group causal transformer with 
curriculum learning. TiBiX~\cite{tibix2024} explores 
bidirectional generation, producing both reports from image 
pairs and synthetic images from reports, using temporal tokens 
to encode visit intervals. MLRG~\cite{mlrg2025} applies 
multi-view contrastive learning with a tokenised absence 
encoding for missing priors. DDaTR~\cite{ddatr2025} captures 
temporal differences through dynamic feature alignment and 
difference-aware residual modules. CXRMate~\cite{cxrmate2024} 
takes a simpler approach, conditioning generation on the prior 
study's report as a prompt. Libra~\cite{libra2025} adds a 
temporal alignment connector to a multimodal LLM for 
paired-image reasoning. All of these methods produce free-text 
reports without spatial grounding. TRACE~\cite{tracev1} is the 
only prior work to combine temporal reasoning with 
bounding-box grounding, but it handles only one finding per 
sample, whereas clinical reports routinely contain multiple 
co-occurring findings.

\paragraph{Region-aware and grounded generation.}
Grounding each finding in the image provides spatial 
explainability and allows radiologists to verify model outputs 
directly. Tanida et al.~\cite{tanida2023rgrg} detect 
anatomical regions with Faster R-CNN and generate per-region 
sentences using a fine-tuned GPT-2. ChEX~\cite{chex2024} 
supports interactive use, accepting textual prompts or 
user-drawn bounding boxes to generate region-specific 
descriptions. MedVersa~\cite{medversa2024} handles detection, 
segmentation, and grounded report generation within a single 
foundation model. MAIRA-2~\cite{bannur2024maira2} generates 
inline bounding-box coordinates for each finding and proposes 
RadFact, an LLM-based framework for evaluating grounded 
factuality. PadChest-GR~\cite{padchestgr2024} contributes a 
bilingual dataset with radiologist-drawn bounding boxes for 
training and evaluation. These grounded methods all operate on 
a single time point and do not capture interval change. Our 
RGCT module takes a different approach: it uses the fixed 
anatomical vocabulary of Chest 
ImaGenome~\cite{wu2021chestimagenome} to pool region-level 
features from a pre-trained temporal encoder, giving the 
language model a per-region change signal alongside spatial 
coordinates.

\section{Dataset}
\label{sec:data}

We construct our dataset from MIMIC-CXR v2.1.0
\cite{johnson2019mimiccxr} and the Chest ImaGenome v1.0 silver
dataset \cite{wu2021chestimagenome}. We use the same sources and the
same official Chest ImaGenome patient-level splits as TRACE
\cite{tracev1}, but address two problems with TRACE's
pair-construction pipeline: (1) AP retakes from a single exam are paired
as if they were a prior/current comparison because they share the
same \texttt{StudyOrder} index, and (2) Chest ImaGenome's NLP-derived
comparison cues fabricate NEW or RESOLVED events whenever a
persistent finding is simply omitted from a follow-up report. The following subsections describe how we construct temporal pairs, identify two systematic errors in the standard pipeline, and detail the cleaning procedure that produces the final benchmark.

\subsection{From Scene Graphs to Temporal Pairs}

\begin{figure}[h]
\centering
\includegraphics[width=0.8\linewidth]{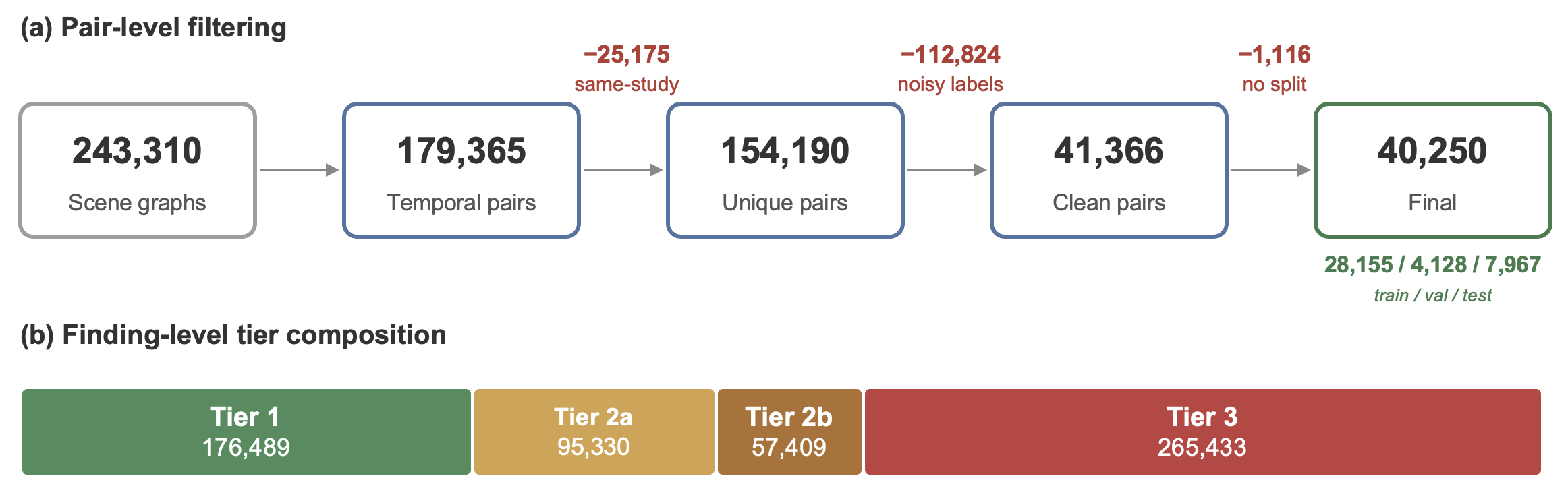}
\caption{Dataset construction pipeline. (a)~Pair-level filtering: 243,310 scene graphs from Chest ImaGenome are paired into 179,365 temporal pairs, then filtered by removing same-study duplicates ($-$25,175), pairs containing any Tier-3 finding ($-$112,824), and pairs outside the official MIMIC-CXR patient splits ($-$1,116), yielding 40,250 pairs (28,155 train / 4,128 val / 7,967 test). (b)~Finding-level tier composition across findings in the 154,190 unique pairs (594,661 findings; 3.9 findings per pair on average): Tier~1 findings (176,489; 29.7\%) have direct NLP comparison cues, Tier~2a (95,330; 16.0\%) are rescued by keyword matching, Tier~2b (57,409; 9.7\%) are rescued by temporal interpolation, and Tier~3 (265,433; 44.6\%) lack evidence and trigger pair removal. A pair is removed if it contains any Tier-3 finding, so the 44.6\% finding-level Tier-3 rate translates to 73.2\% of pairs being dropped.}
\label{fig:data-pipeline}
\end{figure}

For each patient, Chest ImaGenome provides scene graphs with a
\texttt{StudyOrder} index encoding the chronological order of their
studies. Each patient's studies are sorted by this index, and a
temporal pair $(I_p, I_c)$ is formed from every adjacent pair of
studies, so a patient with $n$ studies contributes $n-1$ pairs.
Summed over the 63{,}945 patients with at least two studies, this
construction produces a na\"ive set of 179{,}365 pairs. This is the
construction used by TRACE \cite{tracev1} and, to our knowledge, by
other Chest ImaGenome-based temporal benchmarks.
Figure~\ref{fig:data-pipeline} summarizes the cleaning pipeline:
same-study removal, Tier-3 filtering, and patient-split assignment.
The remaining subsections detail each stage.

\subsection{Issue 1: Same-Study AP Retakes}
\label{sec:data:samestudy}

10.8\% of Chest ImaGenome studies contain two or more images, all of
which are AP retake views (the radiographer repeated the exposure for
quality, not a second exam). These duplicates share a
\texttt{StudyOrder} value, and \texttt{StudyOrder}-based pairing has
no way to tell them apart from a real prior/current pair. When such
a duplicate lands in the middle of a patient's sequence, the na\"ive
pair construction produces a ``temporal'' pair whose two images come
from the same exam, the same minute, and the same report. 25{,}175
pairs, or 14.0\% of the na\"ive set, are of this form.

\begin{figure}[!h]
\centering
 \includegraphics[width=0.6\linewidth]{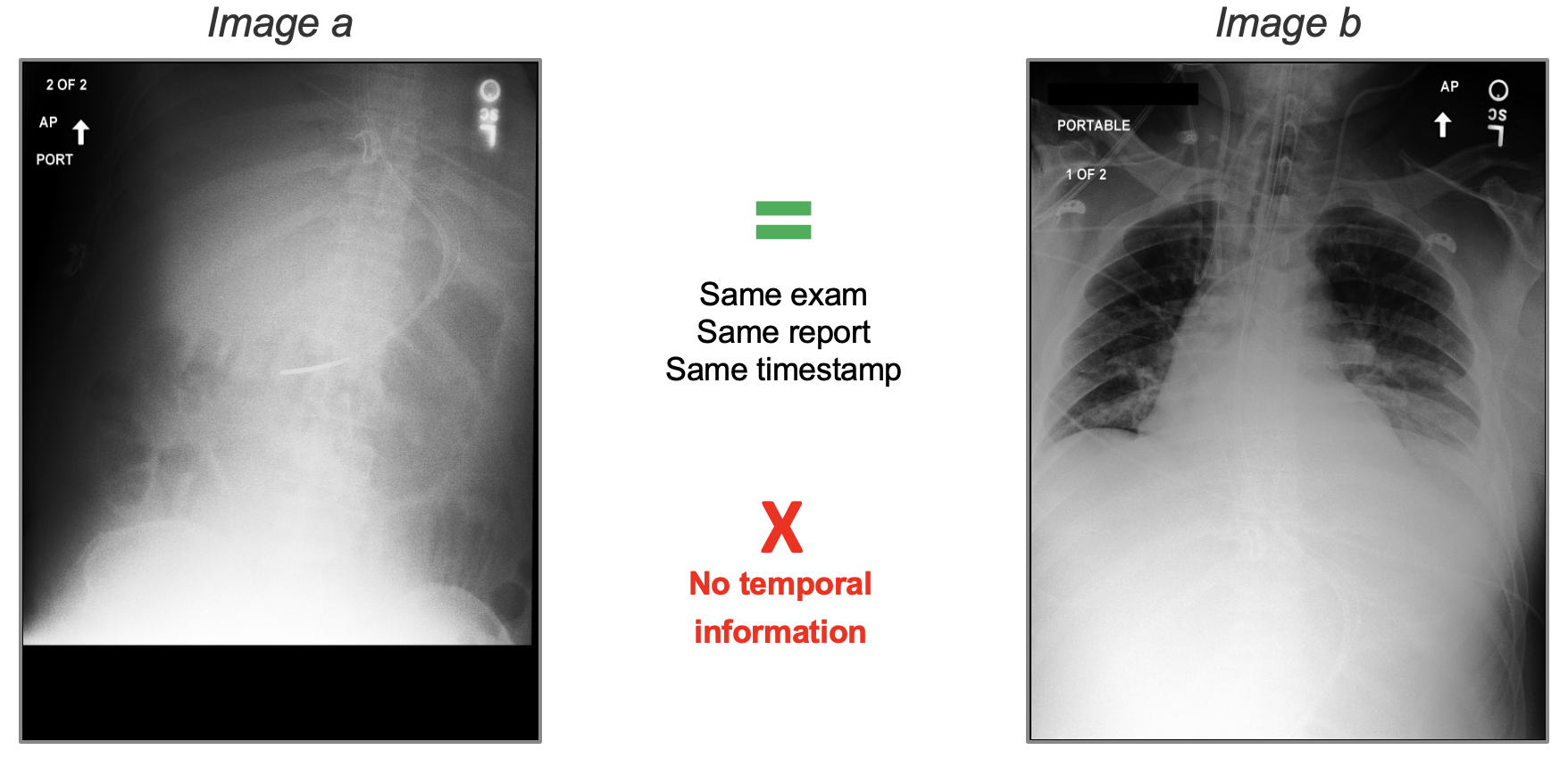}
\caption{A same-study AP retake misidentified as a temporal pair. Both images belong to a single exam (subject \texttt{p10934092}, study \texttt{s54010992}) and share one radiology report. Such pairs carry no temporal information and are removed by requiring $\texttt{prior\_study\_id} \neq \texttt{current\_study\_id}$, eliminating 25{,}175 pairs (14.0\%).}
\label{fig:samestudy}
\end{figure}

A concrete example: subject \texttt{p10934092}, study
\texttt{s54010992}, two AP DICOMs taken at the same timestamp
(Figure~\ref{fig:samestudy}). Both images share one report, which
describes a Dobbhoff tube placement and makes no comparison statement
at all. A model trained on this pair cannot learn anything useful,
because the two scene graphs are identical up to patient positioning.
We filter all such pairs with a one-line check in the pairing loop:\newline
$\texttt{prior\_sg.study\_id} \neq \texttt{curr\_sg.study\_id}$.

\begin{figure*}[!b]
\centering
\includegraphics[width=0.8\linewidth]{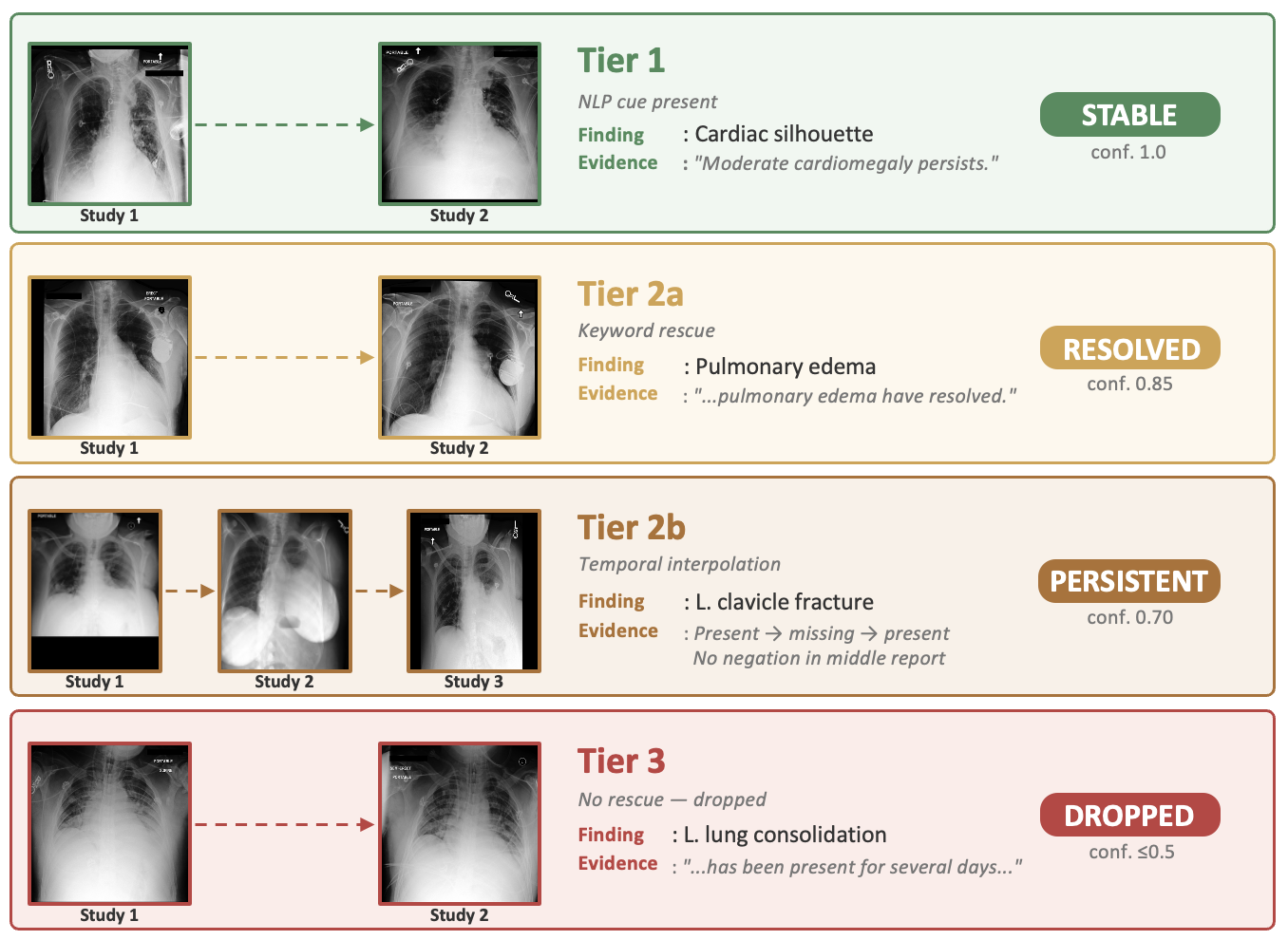}
\caption{One real example per confidence tier. Each row shows the
temporal CXR pair (Study~1 and Study~2), the finding and its
textual evidence, and the pipeline's label decision.
\textbf{Tier~1}: an explicit NLP comparison cue is attached to the
finding.
\textbf{Tier~2a}: the cue is missing, but the keyword
\texttt{resolved} in the current report recovers the label.
\textbf{Tier~2b}: the cue is missing and no keyword is found, but
the finding is present in both the preceding (Study~1) and
following (Study~3) exams with no negation in between, so the gap
is relabeled as \emph{persistent} by temporal interpolation.
\textbf{Tier~3}: no cue, no keyword, and no successful bracketing;
the finding is dropped from supervision.}
\label{fig:tiers}
\end{figure*}

\subsection{Issue 2: Silent NLP Absence, and the Three-Tier Fix}
\label{sec:data:tiers}

Chest ImaGenome attaches NLP-extracted comparison cues to scene-graph
findings, tagging them as \emph{worsened}, \emph{improved}, or
\emph{no change}. Any pipeline that relies solely on these cues
(including TRACE \cite{tracev1}) is implicitly restricted to those
three labels. A finding that appears or disappears from a patient's
scene graphs \emph{without} a cue is silently dropped. We find that
77\% of raw NEW/RESOLVED transitions in Chest ImaGenome are of this
silent kind (computed as the fraction of all finding-level
NEW/RESOLVED transitions that lack any attached comparison cue in the source scene graph). 94{,}117 per-patient gaps exhibit the resulting flicker,
even for physically irreversible findings: the clavicle-fracture gap
rate is 62.4\%, the rib-fracture rate 58.7\%, and the spinal-fracture
rate 50.5\%. Our cleaning pipeline must handle \emph{new} and
\emph{resolved} transitions, so we cannot rely on the cue-only filter.

We classify every finding in every pair into one of three tiers
(Figure~\ref{fig:tiers}):

\begin{description}
\item[Tier 1] (confidence 1.0): a Chest ImaGenome comparison cue is
attached to the finding. The cue directly supplies the label.
\item[Tier 2] (confidence 0.7--0.85): the cue is missing, but we
recover the label from textual evidence. Tier 2a uses keywords
(\texttt{new}, \texttt{resolved}, \texttt{developed}) in the current
report or explicit negation in the prior. Tier 2b uses temporal
interpolation: if a finding is present before the gap and after the
gap with no negation in the middle, the middle must be a silent
miss, and the finding is relabeled as persistent.
\item[Tier 3] (confidence $\leq$ 0.5): a finding appears or
disappears with no textual evidence in either report. We drop these.
\end{description}

Pairs in which every finding is Tier 1 or Tier 2 are kept (41{,}366
pairs, 26.8\% of the post-same-study pool). The remaining 112{,}824
pairs contain at least one unverified Tier-3 finding and are dropped. We validate this decision in
Section~\ref{sec:ablation}: adding the rejected pairs back to
training as weakly-supervised data degrades NLG metrics by up to 15.6\% on the clean test set.

\subsection{Structured Output Format}

\begin{figure*}[!h]
\centering
\includegraphics[width=0.8\linewidth]{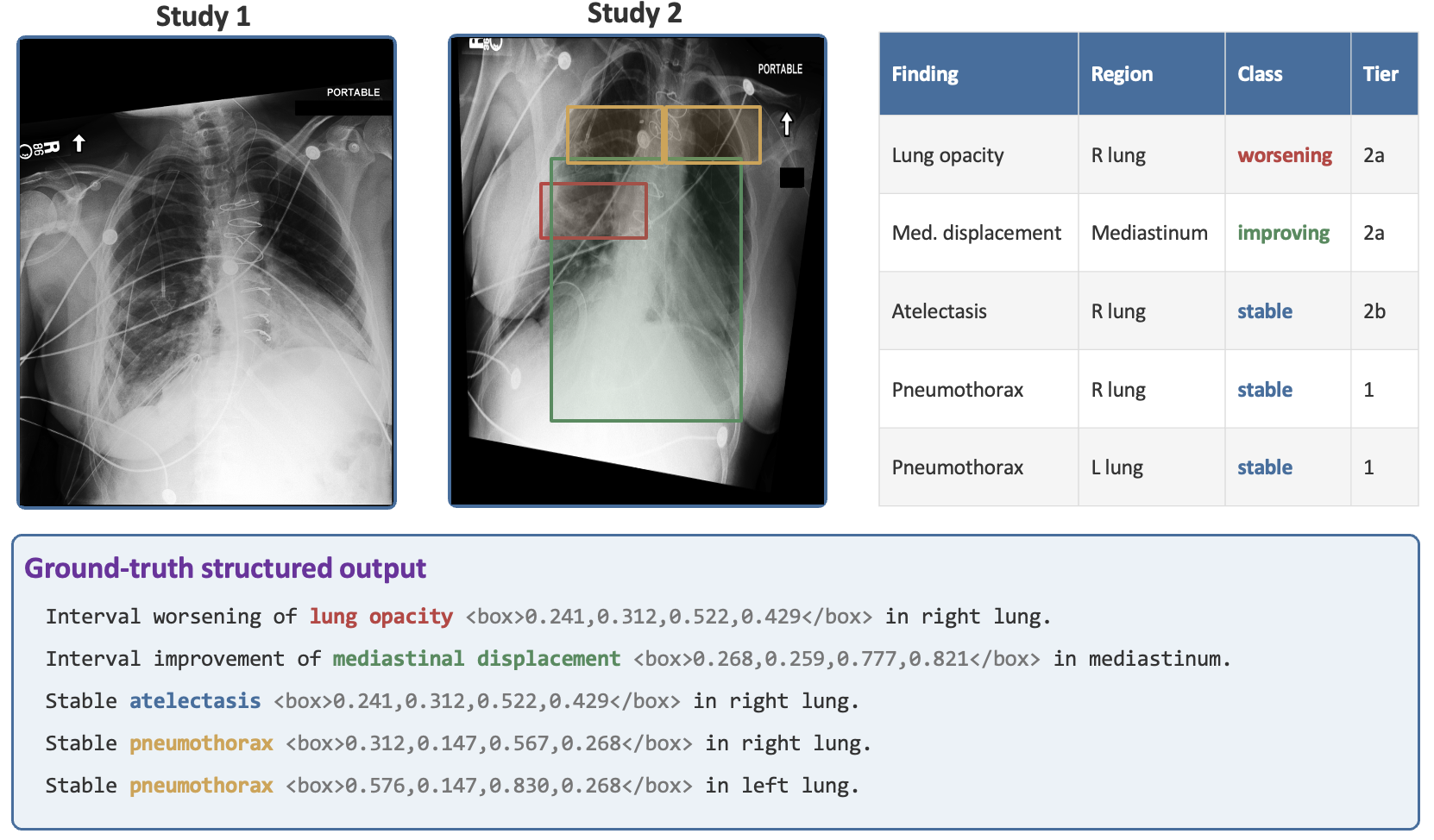}
\caption{A real multi-finding test pair (subject \texttt{p13613806})
that exercises every mechanism of the cleaning pipeline: three
different output classes (worsening, improving, stable), three of
the six underlying change labels (new, resolved, persistent), and
three of the four confidence tiers (Tier~1 direct cue, Tier~2a
keyword, Tier~2b interpolation). Bounding boxes on Study~2 are
color-coded by finding. Boxes are normalized to $[0,1]$; for
\emph{new} findings the box is taken from the current image, for
\emph{resolved} from the prior, otherwise from the current with prior
as fallback.}
\label{fig:multifinding}
\end{figure*}

After cleaning, each pair is annotated with a structured report
consisting of two sections: findings visible in the prior image and
findings in the current image, each tagged with a region name, a
bounding box, and (for the current section) one of three change
classes (stable, worsening, or improvement). Figure~\ref{fig:multifinding} shows a real example from our
test set.

\subsection{Dataset Statistics and Audit}

Of the 41{,}366 clean pairs, we apply the official Chest ImaGenome
patient-level splits and remove 1{,}116 pairs from patients not
in any split or with missing images,
yielding 40{,}250 pairs partitioned into 28{,}155
training, 4{,}128 validation, and 7{,}967 test pairs, covering 41
distinct pathology types. The cleaning pipeline classifies findings 
into six underlying change types (stable, persistent, worsened, 
improved, new, resolved) to determine label reliability. For 
generation, these map to three output classes: \emph{stable} 
(stable or persistent), \emph{worsening} (worsened or new), 
and \emph{improvement} (improved or resolved). 
Table~\ref{tab:class_counts} gives the per-class finding counts 
across splits.

\begin{table}[!h]
\centering
\small
\caption{Per-class finding counts across train, validation, 
and test splits.}
\label{tab:class_counts}
\begin{tabular}{@{}lcccc@{}}
\toprule
Split & Stable & Worsening & Improvement & Total \\
\midrule
Train (28{,}155 pairs) & 68{,}955 & 30{,}733 & 35{,}711 & 135{,}399 \\
Val (4{,}128 pairs)    &  9{,}913 &  4{,}248 &  5{,}455 &  19{,}616 \\
Test (7{,}967 pairs)   & 19{,}594 &  8{,}353 &  9{,}793 &  37{,}740 \\
\midrule
Total (40{,}250 pairs) & 98{,}462 & 43{,}334 & 50{,}959 & 192{,}755 \\
\bottomrule
\end{tabular}
\end{table}

We validate the final splits with an automated audit that checks
for same-study and same-image leaks, bounding-box range violations,
duplicate finding entries, and misplaced temporal labels (\eg
\emph{new} findings in the prior section). All checks pass on all
40{,}250 samples. Patient-level partitioning is inherited from the official Chest ImaGenome splits, ensuring no patient appears in more than one
subset.

\section{Method}
\label{sec:method}

\begin{figure*}[t]
\centering
\includegraphics[width=0.9\textwidth]{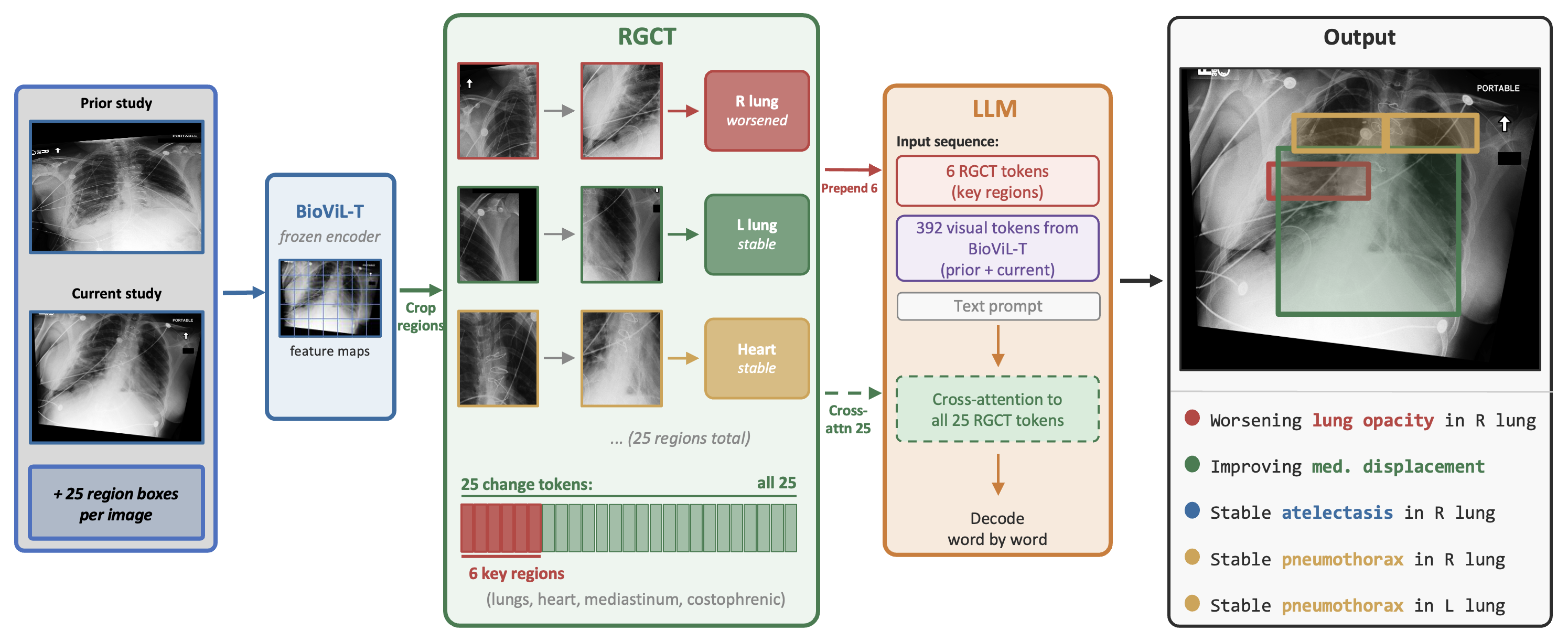}
\caption{Overview of the GRCD architecture. A prior and current
CXR pair are encoded by a frozen BioViL-T encoder into spatial
feature maps. The RGCT module crops 25 anatomical regions from both
feature maps, compares them across time, and produces 25 change
tokens. Six tokens corresponding to key anatomical regions (lungs,
heart, mediastinum, costophrenic angles) are prepended to the LLM
input, while all 25 are injected via gated cross-attention at every
fourth decoder layer. The LLM also receives 392 projected visual
tokens from BioViL-T. The model decodes a structured report with
inline bounding boxes grounding each finding to a region in the
current image.}
\label{fig:architecture}
\end{figure*}

Figure~\ref{fig:architecture} gives an overview of the GRCD
architecture. The model receives a prior CXR $I_p$ and a current CXR $I_c$
along with 25 anatomical bounding boxes from Chest ImaGenome for each
image, and is trained to generate the structured output format defined
in Section~\ref{sec:data}. The pipeline has three components:
(i)~a frozen BioViL-T \cite{bannur2023biovilt} temporal vision encoder,
(ii)~a \emph{Region-Guided Change Token} (RGCT) module, and
(iii)~an autoregressive LLM decoder.
The following subsections describe each component and the training
procedure in detail.

\subsection{Task Formulation}
\label{sec:method:task}

For each pair we form an input sequence
$[\mathbf{T}_{\text{rgct}};\,\mathbf{V}_p;\,\mathbf{V}_c;\,\mathbf{T}_{\text{prompt}}]$,
where $\mathbf{V}_p, \mathbf{V}_c \in \mathbb{R}^{196 \times d}$
($14{\times}14$ spatial grid) are patch tokens projected from the
prior image and the temporally-fused
current image, $\mathbf{T}_{\text{rgct}} \in \mathbb{R}^{6 \times d}$
are six prepended change tokens
(Section~\ref{sec:method:dual}), and $d{=}4096$ is the LLM hidden
size. The model is trained autoregressively on the structured target;
the box delimiters \texttt{<box>}, commas, and digit tokens are
treated as ordinary text and decoded by the LLM head.

\subsection{Region-Guided Change Tokens (RGCT)}
\label{sec:method:rgct}

BioViL-T is a temporal vision encoder that takes both images
simultaneously, so its output for the current image already captures
what changed from the prior. Concretely, BioViL-T encodes
$(I_c, I_p)$ jointly into a feature map
$\mathbf{F}_c \in \mathbb{R}^{512\times14\times14}$ and encodes
$I_p$ alone into $\mathbf{F}_p$ of the same shape.

For each of $R{=}25$ anatomical regions with normalized bounding box
$\mathbf{b}_r$, we crop a region feature by bilinear sampling on a
$2{\times}2$ grid inside $\mathbf{b}_r$ and averaging:
\begin{equation}
\mathbf{f}_r(\mathbf{F}) \;=\; \frac{1}{4}\sum_{i,j=1}^{2}
  \texttt{GridSample}\big(\mathbf{F},\,\mathbf{b}_r\big)_{ij}.
\end{equation}
We then concatenate the prior, current, and difference features for
each region:
\begin{equation}
\mathbf{x}_r \;=\;
\big[\,\mathbf{f}_r(\mathbf{F}_p);\;\mathbf{f}_r(\mathbf{F}_c);\;
       \mathbf{f}_r(\mathbf{F}_c){-}\mathbf{f}_r(\mathbf{F}_p)\,\big]
\;\in\;\mathbb{R}^{1536}.
\end{equation}
The 25 region descriptors pass through an inter-region self-attention
block (8 heads) so that information can flow across regions (e.g.,\ a
worsening left-lung opacity can inform the right-lung token).
We denote the refined descriptors as $\hat{\mathbf{x}}_r$. Each is
then projected to the LLM hidden space and combined with a learnable
embedding that identifies which region it represents:
\begin{equation}
\mathbf{t}_r \;=\; \mathrm{MLP}(\hat{\mathbf{x}}_r) \;+\;
\mathbf{e}_r, \qquad \mathbf{e}_r \in \mathbb{R}^{d}.
\end{equation}
Three auxiliary heads also operate on $\hat{\mathbf{x}}_r$:
(i)~a \emph{change classifier} that predicts a multi-hot label over
\{improved, worsened, stable\} for each region,
(ii)~a binary \emph{finding detector} that predicts whether each
region carries an active finding and also gates the cross-attention
keys and values (Section~\ref{sec:method:dual}), and
(iii)~a \emph{grounding head} that regresses a normalized $[0,1]^4$
bounding box per region. All three provide auxiliary supervision
during training (Section~\ref{sec:method:training}).

\subsection{Dual-Pathway Integration}
\label{sec:method:dual}

The 25 change tokens reach the LLM through two complementary
pathways. 

\textbf{Pathway A} selects the six tokens corresponding to
the six key anatomical regions
$\mathcal{P}=\{ \text{right lung},\, \text{left lung},\,$ 
$\text{cardiac silhouette},\, \text{mediastinum},\, \text{left costophrenic},\, \text{right costophrenic}\}$
and prepends them to the LLM input sequence. Because these tokens
participate in the full self-attention of every decoder layer, they
give the model a direct, coarse-grained spatial signal that is
important for grounding accuracy
(Section~\ref{sec:ablation:integration}).

\textbf{Pathway B} inserts a Flamingo-style \cite{alayrac2022flamingo} gated
cross-attention block at every fourth decoder layer (8 blocks across
32 layers, at indices $\{3,7,\dots,31\}$). Each block normalizes the
hidden states $\mathbf{h}$, attends with 8 heads to the full
25-token bank $\mathbf{T}\in\mathbb{R}^{25\times d}$ scaled by the
finding-detector probabilities $\sigma(\mathbf{d})$, and adds the
result back through a tanh-gated residual:
\begin{equation}
\mathbf{h} \leftarrow \mathbf{h} + \tanh(\gamma)\,W_o\,
  \mathrm{Attn}\!\left(W_q\mathbf{h},\;
                       W_k(\mathbf{T}\!\odot\!\sigma(\mathbf{d})),\;
                       W_v(\mathbf{T}\!\odot\!\sigma(\mathbf{d}))\right),
\end{equation}
where $\gamma$ is a learnable scalar initialized to zero so the block
has no effect at the start of training, and $W_q,W_k,W_v$ use a
256-dimensional bottleneck to keep the parameter count low. The two
pathways are complementary: cross-attention lets every decoder layer
access all 25 regions, while the six prepended tokens provide a
persistent spatial anchor at the input boundary. The ablation in
Section~\ref{sec:ablation:integration} shows that removing either
pathway hurts performance: dropping the prepended tokens reduces
grounding accuracy, and dropping cross-attention reduces text quality.

\subsection{Training}
\label{sec:method:training}

We initialize the vision projector, RGCT encoder, and LoRA adapters
from a cross-attention-only variant trained on the same 28{,}155
pairs (Section~\ref{sec:ablation:integration}), then fine-tune for
5 epochs. LoRA rank is $128$ ($\alpha{=}256$, dropout $0.05$),
applied to all linear layers of the LLM (Vicuna-7B-v1.5
\cite{vicuna2023}); the vision tower stays frozen throughout. We use AdamW with bf16 mixed
precision, weight decay $0$, gradient clipping at $1.0$, and a
cosine learning-rate schedule with 3\% warm-up. We assign separate
learning rates to each parameter group:
$2.5{\times}10^{-4}$ for the LoRA adapters,
$1{\times}10^{-4}$ for the RGCT encoder and the cross-attention
blocks, and $2{\times}10^{-5}$ for the vision-language projector.
The per-device batch size is 32 with no gradient accumulation,
giving an effective global batch size of 128.

The total loss is the sum of the autoregressive language-modeling
loss and three auxiliary terms:
(i)~a focal BCE loss for the change classifier
(focusing parameter $\gamma_f{=}2.0$, weight $0.3$), masked by per-label confidence so
that only Tier-1 and high-confidence Tier-2 labels contribute
gradients;
(ii)~a focal BCE loss for the finding detector (weight $0.5$); and
(iii)~a CIoU loss for the grounding head (weight $0.1$).
All three auxiliary terms drop well below the language-modeling loss
within the first epoch.
\section{Experiments}
\label{sec:exp}

\subsection{Setup and Metrics}
\label{sec:exp:setup}

We evaluate on the 7{,}967 test pairs from the cleaned benchmark
described in Section~\ref{sec:data}. Each test sample contains on average 4.7 findings across multiple regions (higher than the 3.9 pre-cleaning average in Figure~\ref{fig:data-pipeline} because Tier-3 filtering preferentially removes pairs with fewer findings), and 31\% of samples
contain findings with different change labels within the same study.
We generate outputs using greedy decoding and parse the structured
text to extract findings, change labels, regions, and bounding boxes.
Inline \texttt{<box>} coordinates are stripped before computing
clinical metrics. All GRCD results are reported as the mean and
standard deviation over three training runs with different random
seeds.

We measure performance along four dimensions.

\begin{itemize}
\item \textbf{Natural Language Generation}: BLEU-4~\cite{papineni2002bleu},
  METEOR~\cite{banerjee2005meteor}, and
  ROUGE-L~\cite{lin2004rouge}, computed at the sentence level over
  the full structured output.

\item \textbf{Clinical accuracy}: CheXbert
  F1-micro~\cite{smit2020chexbert} and RadGraph
  F1~\cite{jain2021radgraph}, which extract pathology labels and
  medical entities from the generated text and compare them against
  the reference.

\item \textbf{Grounding}: mean Intersection-over-Union (IoU) and
  the fraction of findings with IoU${}>{}$0.5, computed by matching
  each predicted bounding box to the closest reference box.

\item \textbf{Change detection}: report-level accuracy (whether
  the dominant 3-class change label matches) and RGCT region-level
  accuracy (whether the classifier's per-region predictions match the
  ground truth).
\end{itemize}

\paragraph{Comparison with TRACE.}
To ensure a fair comparison, we evaluate the publicly released TRACE
model~\cite{tracev1} on our 7{,}967-sample multi-finding test set
using the same greedy decoding and metric pipeline. 

\paragraph{Retrieval baseline.}
Following Boag~\etal~\cite{boag2020baselines}, we include a nearest-neighbour retrieval baseline: each test image is encoded with the frozen BioViL-T encoder and the training-set report with the highest cosine similarity is copied as the prediction. This
provides a non-learned lower bound; any trained model should
exceed copy-from-training performance.

\subsection{Main Results}
\label{sec:exp:main}

Table~\ref{tab:main} compares GRCD against three baselines, all
evaluated on the same 7{,}967 multi-finding test set: a BioViL-T
classifier (MLP trained on frozen region features), TRACE
(re-evaluated on our data), and a nearest-neighbour retrieval
baseline.

\begin{table}[h]
\centering
\small
\caption{Main results on the 7{,}967 multi-finding test set.
BioViL-T Clf = MLP on frozen BioViL-T region features
(change detection only).
Retrieval = nearest-neighbour by BioViL-T image
embedding~\cite{boag2020baselines}.
TRACE = re-evaluated on our test set~\cite{tracev1}.
GRCD = mean $\pm$ std over 3 seeds.
Best values are \textbf{bolded}.}
\label{tab:main}
\setlength{\tabcolsep}{5pt}
\begin{tabular}{@{}l@{\hspace{60pt}}cccc@{}}
\toprule
Metric & BioViL-T Clf & Retrieval & TRACE & GRCD \\
\midrule
\multicolumn{5}{@{}l}{\emph{Natural language generation}} \\
BLEU-4        & --- & 0.273 & 0.078 & \textbf{0.398}$\pm$0.003 \\
METEOR        & --- & 0.374 & 0.213 & \textbf{0.494}$\pm$0.003 \\
ROUGE-L       & --- & 0.376 & 0.176 & \textbf{0.493}$\pm$0.003 \\
\midrule
\multicolumn{5}{@{}l}{\emph{Clinical accuracy}} \\
CheXbert F1   & --- & 0.368 & 0.175 & \textbf{0.567}$\pm$0.009 \\
RadGraph F1   & --- & 0.353 & 0.222 & \textbf{0.473}$\pm$0.002 \\
\midrule
\multicolumn{5}{@{}l}{\emph{Grounding}} \\
Mean IoU      & --- & --- & \textbf{0.545} & 0.540$\pm$0.005 \\
IoU${>}$0.5   & --- & --- & \textbf{64.0\%} & 61.4$\pm$0.5\% \\
\midrule
\multicolumn{5}{@{}l}{\emph{Change detection}} \\
Report-level     & 37.6\% & --- & 39.8\% & \textbf{53.3}$\pm$0.2\% \\
Balanced acc.$^*$ & 46.6\% & --- & 35.7\% & \textbf{50.0}$\pm$0.3\% \\
\bottomrule
\multicolumn{5}{@{}l}{\scriptsize $^*$Mean per-class recall;
majority baseline (always predict stable) = 33.3\%.}
\end{tabular}
\end{table}

GRCD outperforms all baselines across text and clinical metrics:
BLEU-4 improves from 0.078 (TRACE) and 0.273 (retrieval) to 0.398,
CheXbert F1 from 0.175 / 0.368 to 0.567, and RadGraph F1 from
0.222 / 0.353 to 0.473. The retrieval baseline outperforms TRACE on
NLG metrics, confirming that TRACE's single-finding output format
is mismatched to the multi-finding evaluation.

For change detection, GRCD achieves 53.3\% report-level accuracy
and 50.0\% balanced accuracy, outperforming all three baselines.
The BioViL-T classifier, which operates on the same frozen visual
features but bypasses the LLM, achieves 46.6\% balanced accuracy,
indicating that the visual encoder captures meaningful change
signals but the language model contributes an additional 3.4
percentage points by reasoning over findings in context.

Grounding accuracy is comparable between TRACE and GRCD on the
same test set (IoU${>}$0.5: 64.0\% vs.\ 61.4\%), though TRACE
produces bounding boxes for only 3{,}552 of 7{,}967 samples
while GRCD covers 4{,}161.

\subsection{Per-Class and Per-Region Analysis}
\label{sec:exp:perclass}

Table~\ref{tab:perclass} breaks down change-detection performance
by class and by evaluation level. The RGCT classifier operates on
visual features and produces balanced results across the three
classes, with stable achieving the highest F1 (0.673).

\begin{table}[!h]
\centering
\small
\caption{Per-class change-detection performance. RGCT reports
classifier-level precision, recall, and F1 over 3 classes.
Text reports finding-level accuracy after matching predicted
findings to references by (finding, region).}
\label{tab:perclass}
\begin{tabular}{@{}lcccr@{}}
\toprule
 & \multicolumn{3}{c}{RGCT classifier} & Text \\
\cmidrule(lr){2-4} \cmidrule(l){5-5}
Class & P & R & F1 & Acc. \\
\midrule
Worsened  & 0.588 & 0.404 & 0.479 & 61.2\% \\
Improved  & 0.649 & 0.494 & 0.561 & 52.9\% \\
Stable    & 0.654 & 0.694 & 0.673 & 35.9\% \\
\bottomrule
\end{tabular}
\end{table}

Table~\ref{tab:perregion} shows finding-level accuracy per
anatomical region. The cardiac silhouette is the easiest region
(57.2\%), likely because cardiomegaly is a visually distinct and
frequent finding. The lung regions are the most difficult
(44.9--46.1\%), reflecting the diversity of pathologies that
overlap spatially in the lungs.
Table~\ref{tab:perpathology} stratifies the same accuracy by
pathology type: consolidation (58.4\%) and enlarged cardiac
silhouette (57.1\%) are easiest, while pleural effusion (43.5\%)
and pneumothorax (42.0\%) are hardest.

\begin{table}[!h]
\centering
\small
\caption{Finding-level change-detection accuracy per anatomical
region.}
\label{tab:perregion}
\begin{tabular}{@{}lrr@{}}
\toprule
Region & Findings & Acc. \\
\midrule
Cardiac silhouette  &    880 & 57.2\% \\
Right lung          & 6{,}460 & 46.1\% \\
Left lung           & 6{,}546 & 44.9\% \\
R.\ costophrenic    & 3{,}014 & 44.7\% \\
L.\ costophrenic    & 3{,}085 & 43.7\% \\
\midrule
Overall             & 20{,}007 & 45.6\% \\
\bottomrule
\end{tabular}
\end{table}

\begin{table}[!h]
\centering
\small
\caption{Change-detection accuracy stratified by pathology type.
Only pathologies with $\geq$100 matched findings are shown.
Accuracy is computed on findings correctly matched by
(finding, region).}
\label{tab:perpathology}
\begin{tabular}{@{}lrr@{}}
\toprule
Pathology & Matched & Acc. \\
\midrule
Consolidation                  &    113 & 58.4\% \\
Enlarged cardiac silhouette    &    879 & 57.1\% \\
Atelectasis                    & 2{,}071 & 48.9\% \\
Pulmonary edema                & 1{,}023 & 45.7\% \\
Lung opacity                   & 9{,}451 & 44.7\% \\
Pleural effusion               & 6{,}126 & 43.5\% \\
Pneumothorax                   &    169 & 42.0\% \\
\bottomrule
\end{tabular}
\end{table}

\paragraph{Decomposing the compound metric.}
The finding-level accuracy in Table~\ref{tab:perregion}
is a compound metric: a prediction is counted correct only if the
finding type, anatomical region, \emph{and} change label all match
the reference simultaneously. We decompose it into two stages to
isolate where errors occur.
In the first stage, the model must generate the correct
(finding, region) pair regardless of change label. The model
achieves 53.1\% recall (20{,}041 of 37{,}740 reference findings
are matched) and 57.2\% precision (20{,}041 of 35{,}022 predictions
match a reference).
In the second stage, among the 20{,}041 correctly matched findings,
45.5\% receive the correct change label. Per-class accuracy reveals
that the model detects worsening reliably (61.2\%) and improvement
moderately (52.9\%), but struggles to confirm stability (35.9\%),
where it tends to over-predict change. The low stable accuracy
accounts for most of the gap between the compound metric (45.6\%)
and the model's actual ability to detect directional change.

\subsection{Qualitative Results}
\label{sec:exp:qualitative}

\begin{figure*}[h]
\centering
\includegraphics[width=0.9\textwidth]{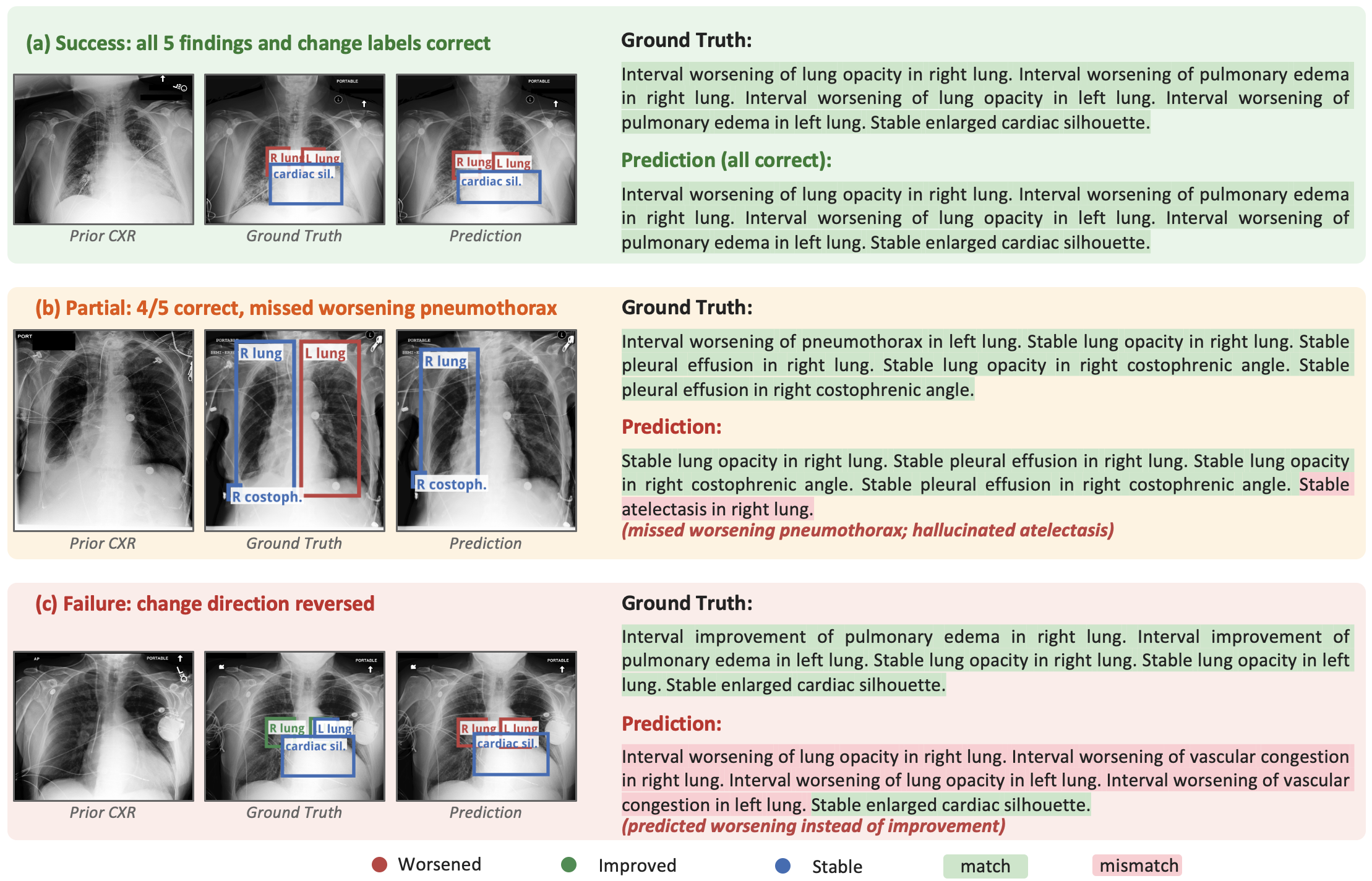}
\caption{Qualitative examples from the test set. Each row shows the
prior CXR, the current CXR with ground-truth boxes, and the current
CXR with predicted boxes. Bounding-box colours indicate change type:
\textcolor{red!70!black}{red} = worsening,
\textcolor{green!50!black}{green} = improving,
\textcolor{blue!70!black}{blue} = stable.
In the report text, \colorbox{green!20}{green highlights} denote
sentences that match the reference and
\colorbox{red!20}{red highlights} denote mismatches.
(a)~All five findings and change labels are correct.
(b)~Four stable findings are correct, but a worsening pneumothorax
is missed and an atelectasis is hallucinated.
(c)~Improving pulmonary edema is predicted as worsening, reversing
the direction of change while localizing to the correct regions.}
\label{fig:qualitative}
\end{figure*}

Figure~\ref{fig:qualitative} shows three test examples that
illustrate the model's strengths and failure modes. In the success
case~(a), GRCD correctly identifies all findings and change labels with well-localized bounding boxes. In the partial-success case~(b), the model correctly localizes stable findings but misses a
worsening pneumothorax, defaulting to stability when directional
change is subtle. In the failure case~(c), the model localizes
findings to the correct regions but reverses the direction of
change. This reversal pattern accounts for a substantial share of
errors, suggesting that distinguishing improvement from worsening remains the primary challenge.

\section{Ablation Studies}
\label{sec:ablation}

Table~\ref{tab:ablation} summarises the ablation experiments.
All variants are evaluated on the same 7{,}967-sample test set;
differences between rows are therefore attributable to the
stated architectural change.

\begin{table}[h]
\centering
\caption{Ablation study. All variants evaluated on the same 7{,}967-sample test set (single seed per configuration; Row~D corresponds to seed~1 of the 3-seed main result).
\textbf{A}: no RGCT (vision + LLM only).
\textbf{B}: 6-region prepend-only.
\textbf{C}: 25-region cross-attention only.
\textbf{D}: full GRCD (dual pathway).}
\label{tab:ablation}
\setlength{\tabcolsep}{3.5pt}
\small
\begin{tabular}{lccccc}
\toprule
Config & BLEU-4 & METEOR & CheXbert
  & IoU${>}$0.5 & ChgDet \\
\midrule
A: No RGCT           & 0.007 & 0.067 & 0.062 & ---   & 37.6 \\
B: 6-reg, prepend    & 0.362 & 0.467 & 0.570 & \textbf{64.2} & 51.6 \\
C: 25-reg, cross-att & 0.394 & 0.476 & 0.512 & 57.9 & \textbf{54.4} \\
D: 25-reg, dual      & \textbf{0.402} & \textbf{0.498}
  & \textbf{0.576} & 61.8 & 53.6 \\
\bottomrule
\end{tabular}
\end{table}

\vspace{-2mm}
\paragraph{RGCT and integration strategy.}
\label{sec:ablation:integration}

Row A removes RGCT entirely, leaving only the frozen BioViL-T
visual tokens and the LLM. Without region-level change information,
the model produces repetitive single-sentence outputs with
near-zero NLG scores and no bounding boxes. This confirms that
RGCT is essential for multi-finding generation: the LLM cannot
learn to produce structured output from global visual features
alone.

Row B adds six macro-region tokens prepended to the input sequence.
This restores full generation ability and achieves the best
grounding accuracy (64.2\% IoU${>}$0.5), because the prepended
tokens give the decoder a strong spatial signal at every layer.

Row C replaces prepending with gated cross-attention over all 25
regions. Text quality improves (BLEU-4: 0.362 to 0.394) and
change detection reaches its peak (54.4\%), but grounding drops
to 57.9\%, indicating that cross-attention alone provides a weaker
spatial signal to the decoder.

Row D combines both pathways. This recovers grounding (+3.9pp
over Row C) while achieving the best text and clinical scores.
The two mechanisms are complementary: prepended tokens supply
direct spatial cues, while cross-attention lets the decoder
selectively query all 25 regions during generation.
The choice between configurations involves a trade-off:
Row~B achieves the best grounding (64.2\% IoU${>}$0.5) while
Row~D achieves the best text and clinical scores. We select
Row~D as the default because multi-finding report quality is
the primary objective.

\vspace{-2mm}
\paragraph{Effect of data cleaning.}
\label{sec:ablation:cleaning}
Training on a larger but noisier superset (the 28{,}155 clean training pairs plus 49{,}717
Tier-3-containing pairs from the training split, i.e.\
the training portion of the 112{,}824 pairs dropped in
Section~\ref{sec:data:tiers}, totalling 77{,}872 pairs) drops METEOR from 0.467 to 0.394 (15.6\% degradation) and ROUGE-L
from 0.449 to 0.416 (7.3\%), even with a two-stage fine-tuning
schedule. Noisy samples teach the model to under-generate findings.
This motivates the strict three-tier cleaning pipeline: retaining
only high-confidence pairs is more valuable than increasing
training-set size with unreliable labels.

\vspace{-2mm}
\paragraph{Number of regions.}
Expanding from 6 macro regions to 25 Chest ImaGenome regions lets
the model ground findings to specific sub-regions (aortic knob,
hilar structures, individual lung zones) not addressable with
coarse tokens. The
benefit is concentrated in text and clinical metrics; the smaller
boxes are individually harder to localise, partly explaining the
IoU gap between Rows B and D.

\section{Discussion}
\label{sec:discussion}

\paragraph{Grounding accuracy.}
On the same multi-finding test set, GRCD and TRACE achieve comparable sample-level grounding (IoU${>}$0.5: 61.4\% vs.\ 64.0\%), though GRCD covers more samples (4{,}161 vs.\ 3{,}552 of 7{,}967).

The high IoU in the original TRACE paper (90.2\%) reflects its single-finding evaluation, where each sample has exactly one predicted and one reference box. In the multi-finding setting, we report the maximum IoU across all predicted and reference box pairs per sample. This metric grows increasingly generous with more findings: samples with a single reference box achieve 51.3\% IoU${>}$0.5, while those with 7{+} boxes reach 70.4\% due to more candidate matches.

A stricter per-finding metric that matches predictions to references within the same anatomical region confirms the true difficulty: region-matched mean IoU drops from 0.444 (single-finding) to 0.195 (7{+} findings), showing that localizing individual findings among overlapping regions is substantially harder than the single-finding case.

\vspace{-2mm}
\paragraph{Data quality over quantity.}
The data-cleaning ablation (Section~\ref{sec:ablation:cleaning})
shows that training on a larger but noisier superset degrades
performance by up to 15.6\% even after a clean fine-tuning stage.
This finding underscores that, for temporal tasks built on silver
annotations, careful curation matters more than scale. The
three-tier pipeline we describe removes 25{,}175 same-study
retakes and retains only pairs whose change labels pass both
all-clean and interpolated-clean checks, reducing the training
set to 28K pairs but yielding consistently stronger models.

\vspace{-2mm}
\paragraph{Limitations.}
The Chest ImaGenome annotations are silver-standard, generated by
an NLP pipeline applied to radiology reports. Despite our three-tier
cleaning, residual label noise persists: 77\% of \emph{new} and
\emph{resolved} labels lack direct textual evidence, and NLP-driven
label flickering affects even clinically irreversible findings.
Additionally, region tokens are defined by rectangular bounding
boxes, which are a coarse approximation of anatomical boundaries;
pixel-level segmentation masks could provide a tighter spatial
signal.

\vspace{-2mm}
\paragraph{Broader impact.}
GRCD is intended as a decision-support tool to assist
radiologists, not to replace them. Automated temporal reports
could reduce dictation time and flag missed changes, but clinical
deployment would require prospective validation, regulatory
approval, and a human-in-the-loop workflow to verify each generated
report before it enters the medical record.

\section{Conclusion}
\label{sec:conclusion}

We presented GRCD, a region-aware framework for grounded temporal
radiology report generation. By combining a region-change encoder
with a dual-pathway integration strategy that both prepends spatial
tokens and applies gated cross-attention, GRCD generates
multi-finding reports with inline bounding boxes grounded to
specific anatomical regions. On a rigorously cleaned subset of
MIMIC-CXR, the model achieves a BLEU-4 of 0.398, CheXbert F1 of
0.567, and RadGraph F1 of 0.473 (averaged over three seeds),
outperforming TRACE on the same test set across all
text and clinical metrics. Ablation experiments confirm that the dual pathway outperforms either integration strategy alone 
on text and clinical metrics, and
that strict data cleaning is essential for silver-annotation tasks.
We hope that the dataset-cleaning methodology and the
dual-pathway architecture provide useful building blocks for future
work on temporal medical-image understanding.

\bibliography{egbib}

\end{document}